%% file: ijcai23.tex
\documentclass{article}
\usepackage{ijcai23}

\usepackage{times}
\usepackage{soul}
\usepackage{url}
\usepackage[hidelinks]{hyperref}
\usepackage[utf8]{inputenc}
\usepackage[small]{caption}
\usepackage{graphicx}
\usepackage{amsmath}
\usepackage{amsfonts}
\usepackage{amsthm}
\usepackage{booktabs}
\usepackage{algorithm}
\usepackage{algorithmic}
\usepackage[switch]{lineno}
\usepackage{booktabs}
\usepackage{multirow}
\usepackage{stfloats}
\usepackage{xcolor}

\newcommand{\shrugtocheck}[1]{#1}

\newcommand{\beginsupplement}{%
        \setcounter{table}{0}
        \renewcommand{\thetable}{S\arabic{table}}%
        \setcounter{figure}{0}
        \renewcommand{\thefigure}{S\arabic{figure}}%
     }

\usepackage{array}
\newcolumntype{L}[1]{>{\raggedright\let\newline\\\arraybackslash\hspace{0pt}}m{#1}}
\newcolumntype{C}[1]{>{\centering\let\newline\\\arraybackslash\hspace{0pt}}m{#1}}
\newcolumntype{R}[1]{>{\raggedleft\let\newline\\\arraybackslash\hspace{0pt}}m{#1}}

\title{ Fairness and representation in satellite-based poverty maps:\\ Evidence of urban-rural disparities and their impacts on downstream policy}

 \author{
 Emily Aiken$^{1, *}$
 \and
 Esther Rolf$^{2, *}$
 \and
 Joshua Blumenstock$^{1}$
 \affiliations
 $^1$UC Berkeley, School of Information\\
 $^2$Harvard University, 
 Data Science Initiative \& Center for Research on Computation and Society \\
 $^*$These authors contributed equally to this work\\
 \emails
 \{emilyaiken, jblumenstock\}@berkeley.edu,
 erolf@seas.harvard.edu
 }

\begin{document}

\maketitle

\begin{abstract}

Poverty maps derived from satellite imagery are increasingly used to inform high-stakes policy decisions, such as the allocation of humanitarian aid and the distribution of government resources. Such poverty maps are typically constructed by training machine learning algorithms on a relatively modest amount of ``ground truth" data from surveys, and then predicting poverty levels in areas where imagery exists but surveys do not. Using survey and satellite data from ten countries, this paper investigates disparities in representation, systematic biases in prediction errors, and fairness concerns in satellite-based poverty mapping across urban and rural lines, and shows how these phenomena affect the validity of policies based on predicted  maps. Our findings highlight the importance of careful error and bias analysis before using satellite-based poverty maps in real-world policy decisions.

    
\end{abstract}

\section{Introduction}

Satellite-based poverty maps are increasingly being used to inform critical policy decisions, 
including estimating interim subnational statistics \cite{hofer2020applying}, targeting humanitarian aid \cite{aiken2022machine,smythe2022geo}, determining eligibility for social services \cite{gentilini2022social}, and estimating the impacts of development programs \cite{huang2021using,ratledge2022using}.
These maps are constructed by applying machine learning (ML) algorithms to high-resolution imagery, based on the premise that the algorithm can learn to predict poverty from pixel data \cite{jean2016combining,yeh2020using,chi2022microestimates}. 
%

However, satellite-based poverty maps are not perfect. 
When poverty predictions exhibit \emph{systematic} errors, their use in policy decisions can lead to disparate and unfair outcomes. 
For example, a program that provides resources
to the regions of a country with lowest predicted wealth might disproportionately ``miss'' poor regions  with substantial infrastructure and large, developed settlements signaling wealth from the sky.
%
In such cases, the use of current satellite-based poverty maps -- which in principle could be used to address the United Nation's (UN) Sustainable Development Goals and other pressing social issues -- might in practice conflict with goals of promoting equity (for example, as formalized in the UN's Leave No One Behind Principle).

The potential for satellite-based poverty maps to aid public policy thus exists alongside the potential for such prediction-based policies to introduce or exacerbate inequities.    
In settings where policymakers may mis-perceive satellite-based maps as technocratic and therefore ``objective" measures of poverty, it is imperative to document 
how systematic errors and biases might arise or compound in satellite-based poverty predictions and their uses in downstream policies.

This paper explores the interconnected phenomena of systematic prediction errors, representation, and unfairness in satellite-based poverty maps, focusing on disparities between urban and rural areas: are satellite-based maps as useful for distinguishing poverty levels within urban and rural areas as between them? Do satellite-based poverty maps tend to overestimate wealth in urban areas relative to rural ones (or vice versa) – and if so, what are the consequences for downstream policy decisions based on such maps? We focus our analyses on urban-rural disparities because (1) previous work has established urban build-up as as a key predictor of poverty in satellite-based machine learning models \cite{yeh2020using,engstrom2017poverty} and (2) many sensitive or protected characteristics – including race, age, and religion – are correlated with
urbanization \cite{ghosh1997changing,kuper2013religion}.

Using survey data and satellite imagery from ten countries (\autoref{table:datasets}), our analysis produces four main results:

First, we document \emph{performance disparities} across rural and urban regions and connect them to potential \emph{representational limitations} of current methods.  
It appears that in many countries, satellite image representations can be used to somewhat accurately differentiate between wealthy and poor regions mainly because these representations capture differences between urban areas (which tend to be wealthy) and rural areas (which tend to be poorer). 
As a result, satellite-based poverty maps are not as effective at differentiating wealth \emph{within} rural and urban parts a country as they are at estimating wealth at a national scale. 

Second, we document nuanced but \emph{systematic biases in prediction errors} for urban and rural areas.
%
In countries where poverty is concentrated in rural areas, predicted wealth in urban areas is under-ranked relative to predicted wealth in rural areas. In contrast, in countries with a high degree of urban poverty, predicted wealth in urban areas is consistently over-ranked relative to predicted wealth in rural areas.
%
%


Third, we study how 
these phenomena
interact to impact the \emph{fairness and effectiveness of downstream policies} based on predicted maps. We simulate hypothetical geographically targeted aid programs which select beneficiary regions using satellite-based poverty predictions.
We observe two contrasting phenomena with opposite effects on selection policies, both tied to the underlying joint distribution of urbanization and wealth. First, systematic over-ranking of rural wealth results in under-allocation of aid to rural areas (particularly when there is a strong correlation between urbanization and ground-truth wealth). Second, overreliance on weaker correlations between urbanization and wealth (arising from representational limitations in satellite imagery) may
result in ``missing'' some of the urban poor. 

Fourth, and finally, we explore options to reduce the exposed disparities in satellite-based poverty mapping. We find that simple recalibration methods can improve predictive accuracy and ameliorate prediction biases in some contexts, but rely heavily on having reliable measures of regions being urban or rural with which to recalibrate. 
%
%


\subsection{Related work}
Satellite-based poverty maps --- which have been studied in the research literature for some time \cite{jean2016combining,yeh2020using,chi2022microestimates,rolf2021generalizable} --- are now being used in real-world policy decisions, including the geographic targeting of social assistance (in Togo \cite{aiken2022machine}, the Democratic Republic of the Congo \cite{gentilini2022social}, and Malawi \cite{paul2021malawi}) and policy impact evaluation (in Uganda \cite{huang2021using} and Rwanda \cite{ratledge2022using}). 
Broad calls to consider fairness and responsibility in satellite-based machine learning -- e.g. in environmental applications \cite{mcgovern2021need}, big data for development \cite{blumenstock2018don}, and remote sensing \cite{burke2021using} --  underscore the importance of evaluating fairness and potential biases in these maps. 

While the implications of algorithmic biases have been documented in settings  from criminal justice \cite{chouldechova2017fairer} and facial recognition \cite{buolamwini2018gender} to credit scoring \cite{liu2018delayed} and resource allocation in healthcare \cite{obermeyer2019dissecting}, they have received relatively little attention in the domain of poverty mapping. 
Recent studies have highlighted specific fairness concerns for particular regions and applications: Kondmann et al. \shortcite{kondmann2021under} investigate statistical bias in estimation of poverty and electrification rates across villages in rural India, Zhang et al. \shortcite{zhang2022segmenting} expose performance gaps of unsupervised transfer learning for landcover classification across rural and urban regions of China, and Smythe and Blumenstock \shortcite{smythe2022geo} evaluate satellite-based poverty targeting in Nigeria. 

However, to date there exists no systematic study of broader fairness concerns in satellite-based poverty mapping --- partly because the data context of low- and middle-income countries (LMICs), where the utility of satellite-derived maps is most distinct, makes it difficult to rigorously evaluate map accuracy and fairness \cite{jerven2013poor,bolliger2017ground,burke2021using,rolf2023evaluation}. Our work builds on previous studies by concretely illustrating how errors and biases in satellite-based poverty maps can translate into disparate outcomes for downstream policy decisions.  

\section{Data and Methods}
Our analysis relies on survey datasets from ten countries matched to featurizations of satellite images. 

\subsection{Survey datasets}
\label{sec:survey_datasets}
We use survey datasets from ten countries in our paper, described in detail in Appendix \ref{app:data} and  \autoref{table:datasets}. In short, we use the following four categories of survey data:

\textbf{Demographic and Health Surveys (DHS)} from Colombia, Honduras, Indonesia, Nigeria, Kenya, the Philippines, and Peru. Each survey was conducted in 2010 or later and interviewed  20,000-60,000 households in 1,000-5,000 clusters. Clusters are small geographic groups of  households, sampled at random or stratified random in each country. Clusters are roughly equivalent to a neighborhood in urban areas (for which the provided cluster centroid is jittered with a 2km radius) or a village in rural areas (for which the jitter is a 5km radius). We use the DHS-constructed asset-based wealth index as the ground truth measure of poverty for each DHS survey, and calculate the average wealth index for each cluster. 

The \textbf{American Community Survey (ACS)} from 2018, which interviewed 1.5 million randomly selected households from all 2,331 Public Use Microdata Areas (``PUMAs") in the United States. We use household income as the ground truth poverty measure in the ACS, and calculate the average household income per PUMA. 

The \textbf{Mexican Intercensal Survey} from 2015, which interviewed 2.8 million households in Mexico’s 2,446 municipalities. We construct an asset-based wealth index from the survey data, using a principle components analysis to project ownership of twelve assets to a unidimensional vector (Appendix \ref{app:data_categorization}). Our ground truth measure of poverty in Mexico is the average asset-based wealth per municipality. 
 
The \textbf{Indian Socio Economic and Caste Census (SECC)} from 2012. We use estimated average per-capita consumption \shrugtocheck{at roughly the village/town level (shrid2) produced by the Socioeconomic High-resolution Rural-Urban Geographic Dataset for India (SHRUG) v2 (an updated version of} \cite{almn2021}) \shrugtocheck{as our reference measure of poverty.} We spatially aggregate small rural shrid2 regions together (Appendix \ref{app:shrug_region_aggregation})
to ensure each observation is a large enough geography and to reduce imbalance between the number of urban and rural regions.
This reduces the number of rural observations from 522,344 to 59,832. There are 3,524 urban regions.

We normalize the poverty values for each country (logged in the US and India\footnote{We log poverty values in the US and India as these values represent consumption distributions, which are right-tailed. In the remaining countries, poverty is measured with asset indices and  we do not use a log transform.})  to zero mean and unit variance. We refer to these poverty measures as ``wealth” throughout. 

Categorizations of each region as either urban or rural are defined by these survey datasets. We refer to these binary labels as ``urbanization'' throughout.

\subsection{Satellite image features}
\label{sec:satellite_imagery_main}
We obtain a set of tabularized features summarizing satellite tiles in each country we study from MOSAIKS \cite{rolf2021generalizable}, accessed via  \url{siml.berkeley.edu} \cite{MOSAIKS_API}. The underlying satellite images are from Planet Labs in 2019.\footnote{Satellite imagery (from 2019) is not obtained from the same time period as all survey datasets, which range from 2010 to 2019. Other work suggests that the impacts of this temporal mismatch are limited \cite{yeh2020using}, and we observe no clear relationship between predictive accuracy and temporal mismatch in \autoref{figure1}.}
Features are generated through an unsupervised machine learning approach based on random convolutional features (RCFs), 
which
are shown to carry skill across a variety of prediction tasks
\cite{rolf2021generalizable}.

RCF embedding functions are essentially a wide and shallow feed-forward convolutional neural network with random but fixed (non-optimized) weights. 
We use RCFs as convenient way to obtain images features with a single, fixed featurization method across countries.
%

The number of tiles per region varies widely between survey datasets: in the DHS, where each cluster has a 2-5km radius, each cluster is represented with 16-88 tiles. In the India, Mexico, and the United States, regions can overlap as few as six tiles or as many as tens of thousands of tiles (\autoref{table:datasets}). For regions that intersect more than 100 tiles, we take a random subset of 100 of the intersecting tiles. We then calculate the average of each MOSAIKS feature for each region, weighted by the overlap between the tiles and the region.

\subsection{Problem formulation and simulation setup}
Our machine learning simulations begin by randomly assigning 75\% of regions in each country to a training set and 25\% to a test set.\footnote{We use uniform random assignment of regions (PUMAs in the US, municipalities in Mexico, aggregated Shrid2 units in India, and clusters in DHS surveys) to train and test sets --- rather than spatial stratification --- as it allows for more consistency across countries, and better reflects the ``in-sample'' scenarios in which satellite-based poverty maps would be deployed \cite{wadoux2021spatial,rolf2023evaluation}.} 
Following Rolf et al. \shortcite{rolf2021generalizable}, in each country we train a ridge regression model to predict average household wealth in training set regions from the associated satellite-derived MOSAIKS features. The objective function is mean squared error, and we tune the $\ell_2$ penalty via three-fold cross-validation on the training set. We then use the trained model to predict wealth for every region in the test set. To account for idiosyncrasies in random train-test splits, we report the mean $\pm$ two std. errors across 100 simulations in all results. 

\subsection{Fairness analysis procedures}
\label{sec:fairness_procedures}
Our analysis focuses on bias and fairness in satellite-based poverty maps along urban-rural lines. First, we document \textit{performance disparities} within and between urban and rural areas, by measuring predictive accuracy (measured with $R^2$ and Spearman’s $\rho$) in the test set overall, in just urban regions, and in just rural regions. Second, we measure \textit{systematic prediction biases} between urban and rural regions when using satellite-based poverty maps, quantified as (1) the mean signed error in wealth prediction for rural and urban areas separately, and (2) the mean error in wealth ranking for rural and urban areas separately. 

We then measure how performance disparities and prediction biases propagate to downstream policy decisions. We simulate hypothetical aid programs using satellite-based poverty predictions to select eligible geographies. To evaluate the implications of performance disparities on simple metrics of \emph{allocational fairness}, we compare the precision and recall (equal by definition in this application \cite{brown2018poor}) of hypothetical programs that target the poorest 20\% of regions in each country as a whole, the poorest 20\% of urban regions, and the poorest 20\% of rural regions. To show how systematic prediction biases propagate to downstream policy decisions in nationwide aid programs, we measure aid allocation (measured as the number of regions selected) to rural areas and urban areas when satellite-based poverty maps are used to select geographies, and compare to allocations when ground truth measures of poverty are used.

\subsection{Recalibration approaches}
\label{sec:methods_calibration}

We explore two recalibration-based options for addressing fairness issues in satellite-based poverty prediction: \textit{mean calibration} (adjusting the means of urban and rural predicted wealth distributions to match the means of the ground truth distributions), and \textit{selection threshold calibration}  (allocating resources to urban and rural areas according to the share of regions that are poor in each group).
For both approaches, we learn the parameters of the calibration procedure on the training set, and apply this learned calibration to the test.
%
%
We investigate whether access to ground-truth urbanization values affects the results our calibration approaches by also attempting calibration with \emph{predicted} urbanization in test regions. 

\begin{figure*}[t!]
\begin{center}
\includegraphics[width=\textwidth]{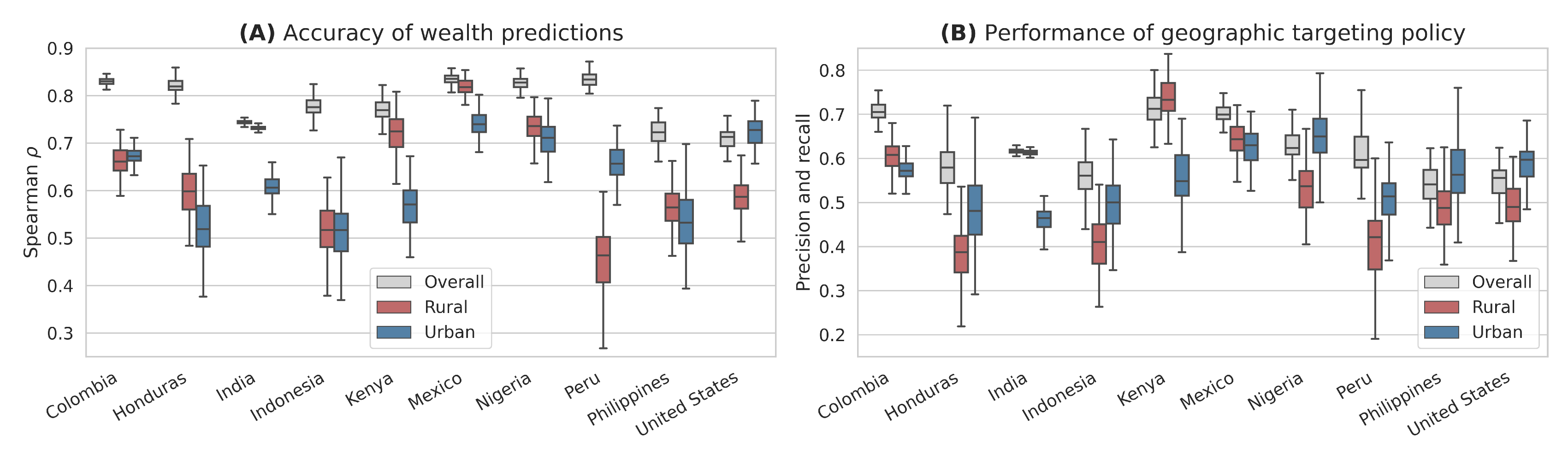}
\end{center}
\caption{\textit{Panel A:} Rank correlation (Spearman’s $\rho$) between predicted and ground-truth wealth are higher in each country as a whole (gray) than within urban (blue) and rural (red) regions in each country. \textit{Panel B:} As a result, an aid program that targets the poorest 20\% of regions in urban (blue) or rural (red) parts of a country has lower accuracy than a program that targets within the entire country (gray). 
}
\label{figure1}
\end{figure*}

\begin{figure*}[t!]
\includegraphics[width=\textwidth]{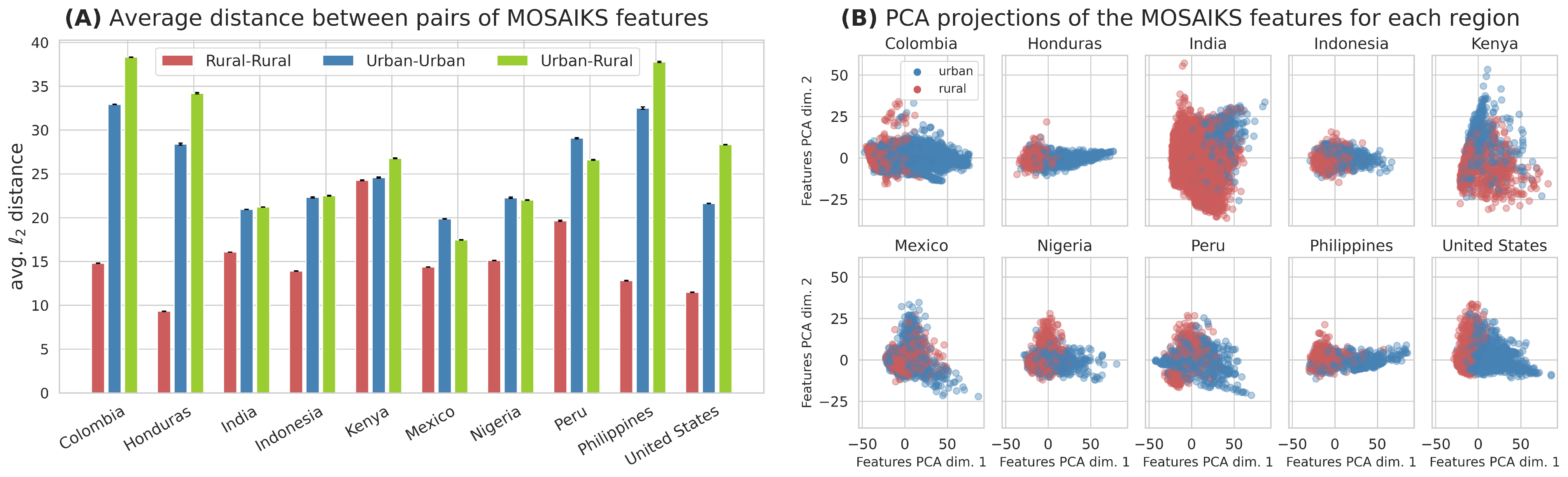}
\caption{
\textit{Panel A:} Average $\ell_2$ distance  between satellite image features for pairs of rural regions, pairs of urban regions, and pairs of urban-rural regions. 
%
%
For India, we randomly sub-sample 2,000 rural and 2,000 urban regions to estimate average distances. 
\textit{Panel B:} Two-dimensional principle components analysis (PCA) projections of the MOSAIKS feature.
Across countries, these dimensions explain between 90.2\%
and 98.5\% of the variation in the 4000 features.
\label{representation_figure}
}
\end{figure*}

\section{Results}
\label{sec:results}
\subsection{Performance disparities and representation}
\label{sec:representational_disparities}

Consistent with past work \cite{yeh2020using,engstrom2017poverty}, we find that satellite-based wealth predictions explain a significant portion of the variance in ground-truth wealth within each of the ten countries we study (mean $R^2$ = 0.47-0.70), and there is a strong correlation between wealth predictions and ground truth (mean Spearman’s $\rho$ = 0.71-0.83). 

In all ten countries, the rank correlation is substantially lower when predictions are evaluated just within urban areas (mean $\rho$ = 0.51-0.74) or just within rural areas (mean $\rho$ = 0.40-0.82) (or both, Figure \ref{figure1}A).This systematically replicates analysis in \cite{yeh2020using} (which documents performance within-urban and within-rural areas for a pooled dataset from several African countries) for ten countries across the globe. There is heterogeneity across countries in terms of which areas are hardest to predict: in three countries (Colombia, Peru, and the United States) predictive accuracy is higher among urban areas than among rural areas, whereas in the remaining seven countries (Honduras, India, Indonesia, Kenya, Mexico, Nigeria, and the Philippines) predictive accuracy is higher among rural areas. In all countries, at least one of urban or rural areas has substantially lower predictive accuracy than the country as a whole (difference in mean $\rho > 0.09$, \autoref{figure1}A).

Why are satellite-based poverty maps consistently worse at differentiating poverty levels within urban or rural areas than within entire countries?
Trends in the imagery and observed wealth data point to the possibility that much of the accuracy observed in country-scale satellite-based poverty maps is due to their ability to distinguish between urban and rural areas.

In each country, there is a strong correlation between the measured (``ground truth") values of wealth and urbanization (\autoref{table:urban}, Spearman's $\rho$ = 0.51-0.77 outside of India and the United States).\footnote{In the India and United States,  $\rho$ = 0.28-0.30. The United States is the only high-income country of the ten we study. The relatively low correlation between wealth and urbanization in India in our data might be due in part to the definition of shrid2 regions, in which many urban regions have large spatial extent while a large majority of region instances are rural (see Appendix \ref{app:data_categorization}). 
} 
We also find that the overall performance of poverty predictions tends to be higher for countries where wealth and urbanization are more correlated (\autoref{figure:correlation_predicts_performance}).\footnote{This trend does not hold, and possibly reverses, when evaluating across only urban or rural regions (also \autoref{figure:correlation_predicts_performance}).}

The potential influence of urbanization can also be seen in the feature representations of the raw imagery --- even before fitting a predictive model ---which already encode high amount of signal as to whether a region is urban or rural (Figure \ref{representation_figure}B). As shown in Figure \ref{representation_figure}A, the average $\ell_2$ distance between features of two rural regions is much lower than that between an urban and a rural region (and two urban regions). We find that a similar overall trend holds when looking at individual MOSAIKS tiles (Figure \ref{fig:feature_distances_1_tile_per_region}), and that satellite imagery is highly predictive of urbanization \ref{table:urban}.

%

Finally, in countries where wealth and urbanizaton have a strong correlation, the differences between the predictive accuracy of satellite-based wealth predictions and satellite-based predictions of a region being \textit{urban} are small (mean difference in Spearman's $\rho$ = 0.07-0.26 outside of the United States and India, \autoref{table:urban} and \autoref{figure:urbanwealth}). Along with the results in Figure \ref{representation_figure}, the close relationship between predicting urbanization and predicting wealth from satellite imagery hints at potential concerns about representations of poverty in satellite imagery akin to stereotype bias \cite{abbasi2019fairness,boyarskaya2020overcoming}, a particular type of representational harm in which the observed data on individuals in a group are more closely related than a more comprehensive characterization of those individuals would warrant.  

Taken together, these results suggest that representations of poverty in satellite imagery beyond urbanization are present but often limited. 
As such, a concern for policy is that 
applications that “zoom in” on urban or rural areas (for example, calculating interim subregional poverty statistics or running an aid program in just urban or rural areas), predictive accuracy for identifying poverty from satellite imagery --- and the accuracy of downstream decisions --- is likely to be substantially lower than 
an overall accuracy estimate would suggest.


\begin{figure*}
\includegraphics[width=\textwidth]{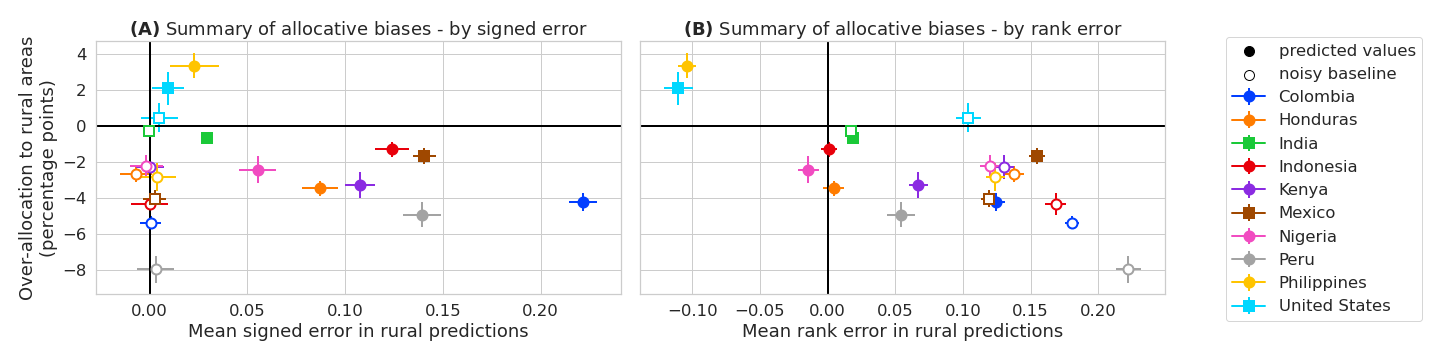}
\caption{Trends in allocative bias in using satellite-based poverty predictions to allocate aid. \textit{Panel A}: Over-allocation of aid to rural areas vs. mean signed error in poverty prediction in rural areas. \textit{Panel B}: Over-allocation of aid to rural areas when  vs. mean rank error in prediction in rural areas. Filled in markers show biases in satellite-based predictions; faded markers show the noised wealth baseline.}
\label{figure2b}
\end{figure*}

\subsection{Systematic biases in prediction errors }
\label{sec:prediction_biases}
In light of the limitations to poverty representations in satellite imagery, a further concern for satellite-based poverty mapping is possible systematic biases in prediction errors. 

We begin by documenting mean \emph{signed errors} in predictions, finding that across countries, wealth in urban areas is under-predicted and wealth in rural areas is over-predicted (\autoref{figure2b}A,   \autoref{figure:allocation}). This phenomenon may simply reflect a statistical bias toward the mean prediction -- in all countries urban areas are on average richer than rural areas (\autoref{table:urban}). 

The mean error in \emph{wealth ranking} across countries exhibits biased errors in both directions:
in Nigeria, the Philippines, and the United States, rural areas are under-ranked by wealth predictions; in Colombia, India, Kenya, Mexico, and Peru, rural areas are over-ranked; and in Honduras and Indonesia, there is no statistically significant difference in ranking between urban and rural areas (\autoref{figure2b}B). 

An important question is whether these same biases could arise if simply using a lower-quality wealth label, rather than satellite-based predictions.
Figures \ref{figure2b} and \ref{figure:allocation} therefore include \textit{noised-wealth baselines}, in which we add Gaussian noise to the ground-truth wealth labels with zero mean and isotropic covariance calibrated to the mean squared error of the satellite-based predictions. 
This allows us to test whether prediction biases of satellite-based models are systematically different than those that would be observed under a  model of independent, additive prediction noise.
Since urban areas have higher average wealth than rural areas across countries in our study, 
we expect the noised income baseline will over-rank rural wealth and under-rank urban wealth. 

Both the satellite-based poverty predictions and the noised-income baseline over-rank wealth in rural areas in most countries (horizontal axis of  \autoref{figure2b}B). The degree of over-ranking tends to be higher for the noised baseline than the satellite-based predictions.
The notable exceptions are the United States and the Philippines, where prediction biases from satellite imagery run in the opposite direction of those from the noised wealth baseline (wealth is under-ranked in rural areas by satellite-based predictions and consistently over-ranked by the noised wealth baseline in these two countries). 
%
We explore possible drivers of these differences in Section \ref{sec:drivers_of_unfairness}.

\subsection{Implications for downstream policies} 
\label{sec:implications_for_policies}
To study 
the extent to which performance disparities and systematic prediction biases can propagate to allocative unfairness in downstream policy decisions, we simulate hypothetical geographically targeted aid programs in each country, as described in Section \ref{sec:fairness_procedures}.

\textbf{Geographic targeting effectiveness.}
We find that the disparities in predictive performance between urban and rural areas documented in Section \ref{sec:representational_disparities} reduce the effectiveness of downstream decisions made using the satellite-based poverty predictions. 
A simulated social protection program aiming to select the poorest 20\% of regions nationwide using satellite-based poverty maps tends to have relatively high recall and precision (54-71\%), whereas programs identifying the poorest 20\% of regions \textit{within} urban or rural areas have lower recall and precision (38-73\% in rural areas and 46-65\% in urban areas, \autoref{figure1}B). 

\textbf{Allocative unfairness.}
%
The systematic biases in ranking of poverty by satellite-based predictions (Section \ref{sec:prediction_biases}) suggests a risk of allocative unfairness when using satellite-based poverty predictions to inform policy.
In our simulated nationwide aid programs, in countries where the relationship between urbanization and wealth is strong (Colombia, Honduras, India, Indonesia, Kenya, Mexico, Nigeria, and Peru), aid tends to be under-allocated to rural areas (by 1-5 percentage points) compared to what would be allocated using ground truth wealth from the survey data. 
In countries where correlation between urbanization and wealth is weaker (the Philippines and the United States), aid tends to be over-allocated to rural areas (by 2-3 percentage points, \autoref{figure2b} and \autoref{figure:allocation}). This latter pattern 
runs in the opposite direction for the noised wealth baseline (
faded markers in \autoref{figure2b}), indicating that error structures specific to satellite-based wealth predictions are driving allocative unfairness, rather than general degradation of wealth estimates.

\begin{figure*}
\centering
\includegraphics[width=.85\textwidth]{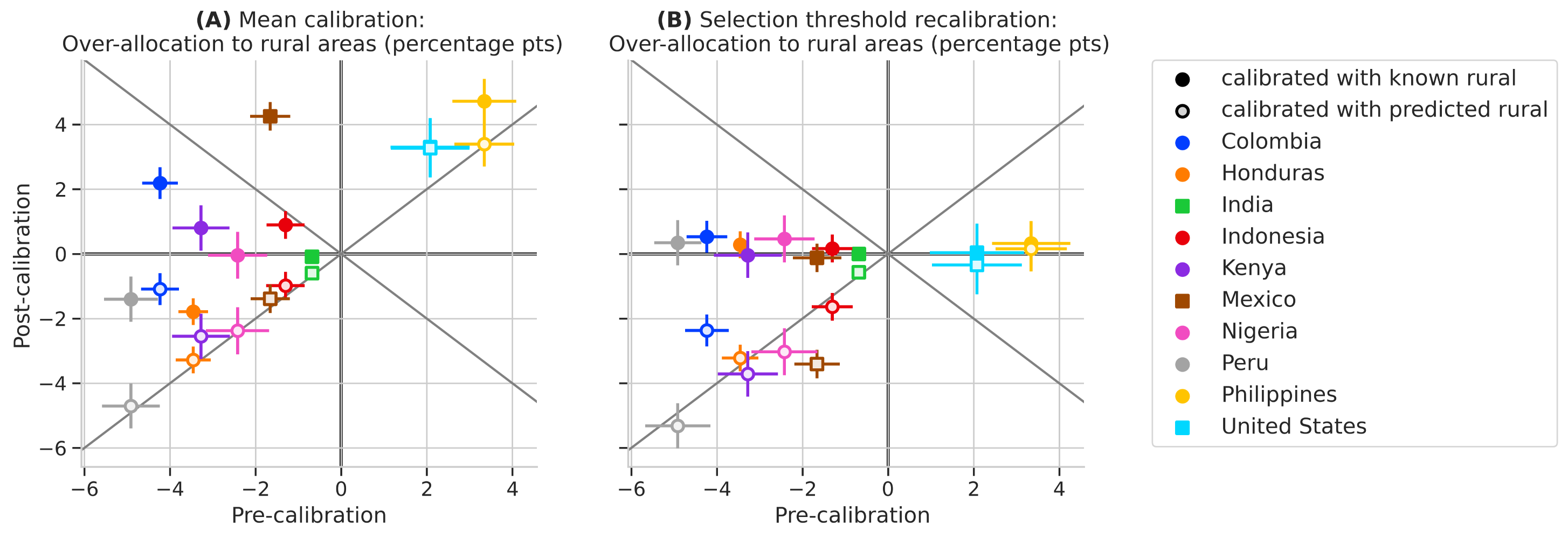}
\caption{Two recalibration options. \textit{Panel A}: Difference in allocation rates to rural regions (between using predictions and survey data to assign allocations), for predictions with and without calibration, using the mean calibration strategy. 
\textit{Panel B}: The same for selection threshold calibration strategy.
A difference of $0$ indicates an exact match with allocations based on survey data.
Filled in markers show correction with known values for urban/rural in satellite-based predictions; empty markers show the result of recalibrating with predicted urban/rural values.
}
\label{figure3}
\end{figure*}

\section{Investigating drivers of allocative unfairness}
\label{sec:drivers_of_unfairness}

The nuanced patterns of allocative unfairness in Section \ref{sec:implications_for_policies} can be at least partially explained by characterizing two phenomena driving errors in satellite-based predictions and ranking of wealth between urban and rural areas: 

\textbf{Reversion towards the (sample) mean.} One possible driver of allocative unfairness is that 
predicted wealth can be biased upward for low wealth regions and downwards for high wealth regions, towards the overall mean wealth value in the training data (as described in Section \ref{sec:prediction_biases}). 
In our simulated aid program, the upward bias of wealth rankings in rural areas results in under-allocation of aid to rural areas. Colombia, Honduras, India, Indonesia, Mexico, Nigeria, and Peru are all emblematic of this pattern to varying degrees. 
Notably, allocative biases are \textit{less} severe for many of these countries with satellite-based errors than would be expected with classical Gaussian prediction errors (simulated with the noised wealth baseline in \autoref{figure2b}). 
One possible explanation for this pattern is that a second driver of allocative unfairness in satellite-based poverty predictions --- described below --- works in the opposite direction of classical prediction error.



\textbf{Reliance on correlations between urbanization and wealth.} A second potential driver of allocative unfairness is a limited predictive power beyond identifying built-up areas (established in Section \ref{sec:representational_disparities}). 
If variation in predicted wealth is driven by urbanization, whereas variation in true wealth is driven by more factors, satellite-based poverty prediction algorithms might “miss” populations of urban poor, having associated them with urbanized regions tending to be wealthy.
The United States and the Philippines – which have the lowest and third-lowest correlation between urbanization and wealth of all the countries we study, and the lowest overall prediction performance (\autoref{table:urban}) – demonstrate this pattern. 

While these two phenomena have different effects on the allocation rate to urban and rural areas, it is possible (and likely) for them to manifest jointly.\footnote{We discuss this issue further and propose summary statistics to help measure causes of each driver in \autoref{app:supplement}.} Summarized in \autoref{figure2b}, for most countries the first driver seems to have the dominant effect on allocation rates, excluding the United States and the Philippines, where the allocative differences appear to be driven mostly by the second phenomenon.

\section{Addressing allocative unfairness}
\label{sec:recalibration}

We test two approaches to addressing the issues of allocative unfairness characterized in Section \ref{sec:implications_for_policies}.
%
%

First, when we know which regions are classified as urban or rural, 
we can recalibrate the prediction distributions within urban and rural areas to align with the true per-group means in the training data.
This addresses the “reversion to the mean” phenomenon in an application-agnostic way. 
We refer to this procedure as \emph{mean recalibration}, and implement it by learning an additive offset for each group 
so that the predicted mean in each group matches the true group mean.


A second option is to directly address allocational unfairness in the context of resource allocation 
by setting different eligibility thresholds for urban and rural regions. We refer to this option as \emph{selection threshold calibration}, and implement it by setting per-group allocation thresholds to match the fraction of allocations that would be sent to urban and rural areas using the reference wealth label values of the training set. 

%

\subsubsection{Mean calibration}

\autoref{figure3} shows that applying mean calibration often produces downstream allocations that are closer to allocations based on ground-truth wealth measures. 
Mean calibration successfully reduces systematic prediction bias across urban and rural areas, and even slightly increases population level performance for some countries (increase in $R^2$ of 0.00 - 0.02, increase in Spearman $\rho$ of 0.00-0.02; \autoref{figures1}).

However, there are two important caveats to the  mean calibration strategy. 
First, it only addresses the first driver of unfairness in Section \ref{sec:drivers_of_unfairness} --- reversion towards the mean. Across  countries, mean recalibration increases the allocation to rural regions (evidenced by points above the $y=x$ line in \autoref{figure3}) due to the increased separation between predicted wealth of rural and urban regions. In countries where the dominant trend affecting allocation rates is missing the urban poor (the Philippines and the United States), deploying this recalibration strategy can exacerbate allocative differences. For simulations in Mexico, mean recalibration also introduces an allocative bias toward over-targeting rural regions that was not present in the original uncalibrated predictions. 

Second, this simple mean recalibration strategy works only when ground truth labels for being urban or rural are known everywhere (that is, everywhere that the satellite-based poverty map will be used --- not just in the training set). When we use satellite-based predictions for whether a region is urban or rural 
to perform mean recalibration in the test set, allocative bias is not significantly improved in most countries (non-filled-in points 
in \autoref{figure3}A). 

\subsubsection{Selection threshold calibration}
 
When using ground truth indicators of urbanization, 
threshold calibration results in 
allocations that are close to what would be allocated with knowledge of true wealth values (confidence intervals for filled-in points in \autoref{figure3}B all overlap the $y=0$ line). 
This should be expected in our experimental setup, so long as the distributions of urban and rural wealth   in the training set match those in the test set.

When satellite-based predictions for urbanization are used to perform selection threshold calibration in the test set, allocative bias is not improved --- the same pattern observed in mean recalibration. 
It is possible that since wealth predictions and urban build-up predictions are closely related (Section \ref{sec:representational_disparities}), there is little additional signal in urban build-up predictions that is useful for calibration.

\section{Discussion}

Our work raises and investigates  two main concerns relevant to researchers  and policymakers interested in building and deploying satellite-based poverty maps for policymaking.

First, \textbf{there are \textit{performance disparities} in predictive accuracy for identifying wealth levels within urban and rural areas in comparison to between them, explained partly by somewhat \emph{limited representations} of poverty in satellite imagery beyond urbanization}. In particular, wealth is better differentiated between urban and rural areas than within urban or rural parts of a country (Figure 1A). Simulated aid programs that target only urban or only rural areas have lower recall than national-scale programs that can leverage the differences in urban and rural wealth (Figure 1B).

The main implication of this result for real-world deployments is that while satellite-based poverty programming at a country scale may be relatively accurate (as documented in past work \cite{jean2016combining,yeh2020using,chi2022microestimates}),  effectiveness may be substantially lower if programs are deployed just for urban or rural areas (as is fairly common in anti-poverty programming \cite{lindert2020sourcebook}). 


For researchers in machine learning, our results suggest that a focus on building predictive models that represent and distinguish wealth levels within urban and rural areas will be essential for making satellite-based poverty maps a useful and fair measurement tool. Other digital data sources, such as mobile phone data \cite{blumenstock2015predicting,steele2017mapping}, social media data \cite{fatehkia2020mapping,chi2022microestimates}, or information from crowdsourced maps \cite{tingzon2019mapping} may be helpful for improving representation and within-urban and within-rural differentiation. 

Our second main finding is that \textbf{\emph{systematic prediction biases} in poverty predictions between urban and rural areas can result in \emph{allocative bias} in downstream policy decisions.} 
The direction of prediction biases and downstream disparities in allocations depends on the underlying joint distribution of poverty and urbanization: satellite-based poverty maps may ``miss" populations of urban poor in countries with pockets of urban poverty, whereas in countries where poverty is concentrated in rural areas, policies based on satellite-based poverty maps are likely to over-allocate aid to urban areas. The main implication of this result for policymakers is that urban-rural biases may be present even in national-scale policies using satellite-based poverty maps, and such maps should always be audited for bias before deployment. 

We test two simple yet promising approaches to addressing systematic prediction biases through recalibrating predictions or selection thresholds, but
both rely on having access to ground-truth labels for regions being urban or rural in all areas where the map is deployed.  
Imputed urban/rural values are available at an increasingly high resolution globally 
\cite{rao2015small}; 
evaluating whether such estimates are sufficient for model recalibration will be an important topic for future work. More generally, more sophisticated statistical approaches to addressing prediction bias may improve upon the ones we propose here \cite{proctor2023parameter}. 

The real-world implications of performance disparities and prediction biases for downstream analyses and policies are likely to be multi-faceted. 
We study in detail the implications for one downstream use of satellite-based poverty maps: the geographic targeting of humanitarian aid. 
A similar analysis could be applied to understand
implications of disparities and biases for other uses of satellite-based predictions, such as the estimation of sub-national statistics \cite{hofer2020applying,NBERw31044} and causal inference on the effects of anti-poverty programs \cite{huang2021using,ratledge2022using}.
%

%
%

In summary, we find consistent evidence of disparities in satellite-based poverty maps across ten countries, with different social structures, time scales, and modes of ground truth data collection.
An important complementary analysis, however, would seek to understand how the disparities we identify interact within a single complex sociopolitical context. For example, we studied disparities only across urban and rural areas; developing a more comprehensive set of concerns will crucially rely on local settings of model use.
Such context-driven work, along with the empirical results presented here, can help policymakers realize the potential of satellite-based poverty mapping while mitigating the risk that such maps introduce bias or amplify existing inequities. 



\clearpage

\bibliographystyle{named}
\bibliography{ijcai23}

\section*{Ethical Statement}
This paper seeks to expose and quantify a potentially critical ethical issue in satellite-based poverty prediction: issues of fairness within and between urban and rural areas. However, our work here still sits squarely within computational and algorithmic aspects of fairness. 
%
%
By focusing on trends across ten very different countries,
the analysis in this paper is largely devoid of the full social context of poverty mapping and aid allocation in the each individual country we study. Country-specific and human-centered work on local conceptions of fairness in such policies will complement the analysis in this paper.


\section*{Acknowledgments}
Aiken acknowledges support from a Microsoft Research PhD Fellowship. Blumenstock acknowledges support from the National Science Foundation under CAREER Grant IIS-1942702.
Rolf acknowledges support from the Harvard Data Science Initiative and the Center for Research on Computation and Society.

We thank Paul Novosad and Sam Asher for sharing with us with an early release of the SHRUG v2 dataset, and for feedback on an earlier draft of this work. We thank Gabriel Cadamuro, Tamma Carleton, Guanghua Chi, and Jonathan Proctor for helpful feedback on the paper. 

 \clearpage

   \appendix
   \onecolumn
   \beginsupplement
  \section{Appendix: Data details}
   \label{app:data}
   \input{supplement_data}

  \clearpage
  \section{Appendix:  Supplementary analysis, figures, and tables}
  \label{app:supplement}

\input{supplement_results}

\end{document}

%% file: supplement_data.tex
\subsection{Categorization of rural vs. urban, calculation of wealth index}
\label{app:data_categorization}
\paragraph{Demographic and Health (DHS) surveys.}
Each DHS survey uses a country-specific rule to define which areas are rural and which are urban; rural ratios range from 30\% in Colombia to 64\% in the Philippines (Table \ref{table:datasets}).

\paragraph{American Community Survey (ACS).}
Categorizations of PUMAs as urban or rural are from Murray \shortcite{murray2022data}; 42\% of PUMAs are categorized as rural. Rural PUMAs are defined by the ACS as ``an agricultural or otherwise sparsely populated PUMA a largest place of fewer than 20,000 people that is not contiguous with another place."

\paragraph{Mexican Intercensal Survey.}
As discussed in Section \ref{sec:survey_datasets}, we construct an asset-based wealth index from the Mexican Intercensal data, using a principle components analysis to project ownership of the twelve assets (electricity, landline phone, mobile phone, internet, car, hot water, air conditioning, computer, washing machine, refrigerator, TV, and radio) to a unidimensional vector. The asset index explains 34\% in the variance in ownership of the underlying assets. Our ground truth measure of poverty is average asset-based wealth per municipality. Municipalities are assigned to urban or rural according to the Mexican government's defnition of rurality (recording in the intercensal survey): rural municipalities are those where the majority of the population lives in communities of less than 2,500 people; the remaining municipalities are urban \cite{bada2022persistent}. 

\paragraph{Indian Socio Economic and Caste Census (SECC) (via SHRUG).}

Shrid2s are geographic units defined and used in the SHRUG database to be consistent with census region definitions over time in India. As discussed in the documentation of shrids for v1.5\footnote{\url{https://shrug-assets-ddl.s3.amazonaws.com/static/main/assets/other/shrug-codebook.pdf}}, each shrid2 region (shrid units for v2 of the dataset) can thus contain multiple villages or towns. The small-area estimation method for computing per-capita consumption estimates for each shrid unit is described in \cite{almn2021}.

Rural and urban units are defined by the 2012 SECC data in the SHRUG database, which separates per capita consumption estimates by urban and rural. Of the 525,868 original shrid2 units, 126 (0.024\%) have SECC values for both rural and urban consumption. We categorize these units as be urban, taking rural regions to be those only with rural consumption. For the 126 units with both urban and rural consumption, we calculate total consumption as a weighted average of urban and rural consumption in the shrid, weighting by the urban and rural population from the 2011 Indian population censuses (also aggregated to shrids as in \cite{almn2021}).

\subsection{Combining rural SHRUG regions to larger geographical extents}
\label{app:shrug_region_aggregation}

In the original SHRUG v2 dataset, there are 3,524 urban (or both urban and rural) shrid2 units and 522,344 rural units.
Because some shrid2 regions are very small in geographic extent, we combine rural shrid2 to reduce the imbalance between the number of urban and rural observations. 
This also ensures that several MOSAIKS tiles overlap with each cell for most observation units. Only rural shrid2s are merged together; we do not alter the extents of urban shrid2s.

We use the following procedure to merge small rural shrids within the district administrative level (1 level finer than states). 
Within each district, we iteratively find the smallest remaining rural extent (by area). We merge this extent with a neighboring geometry according to the following rules:
(1) only neighboring geometries currently made up of fewer than 25 regions are eligible to be merged, (2) of the candidate neighbors, the one with the highest boundary overlap with the district to be merged is chosen.
If there are no feasible neighbors to merge with, the  geometry will stay as is and be removed from the mergeable list.
We repeat this process until there are no more  mergeable geometries (``mergeable'' meaning geometries of area less than 25km$^2$) with at least one neighbor that satisfies rule (1) above).

When geometries are merged according to this process, per capita consumption estimates are computed as a weighted average of the per capita consumption estimates at shrid2 level, where weights in the average are proportional to the 2011 Indian census population counts for each shrid2.

Before this procedure, there were $3,524$ urban geometries (median area  13.9km$^2$) and $522,344$ rural geometries (median areas 2.92km$^2$). After aggregation, there are $3,524$ (median area  13.9km$^2$) and $59,832$ rural geometries (median areas 40.8km$^2$).

\subsection{MOSAIKS features}
\label{app:mosaiks_features}
As mentioned in Section \ref{sec:representational_disparities}, for each instance (region), the random convolutional feature (RCF) representation is an average of up to 100 MOSAIKS tiles overlapping with the geographic extent of the region.
The minimum, maximum, and average number of MOSAIKS tiles per region in each country is given in \autoref{table:datasets}. 

Figure \ref{representation_figure}A plots the average Euclidean distances between image features for pairs of regions, where the pairs considered are: both rural regions, both urban regions, and one rural one urban region.
In most countries we study, feature representations of urban regions tend to be more similar to features of other urban regions than they are to features of rural instances. 
Rural regions tend to be closer to each other in feature space than urban regions are to each other, though this could be partly due to features in rural regions being an average over more MOSAIKS tiles on average than features corresponding to urban regions.  
Thus, it is difficult to say how differences in average feature distances measured in \ref{representation_figure}A affect performance differnces across urban and rural regions (\autoref{figure1}).
On the one hand, the smaller variation of feature representation within rural regions could contribute to lower predictive performance for rural regions. 
On the other hand, when this lower variation is due to averaging of MOSAIKS more tiles across the rural geographies than urban geographies, the feature representation may be in some sense more precise for rural regions, which could contribute to higher predictive performance in rural regions.

\begin{figure}[t!]
\begin{center}
\includegraphics[width=.7\textwidth]{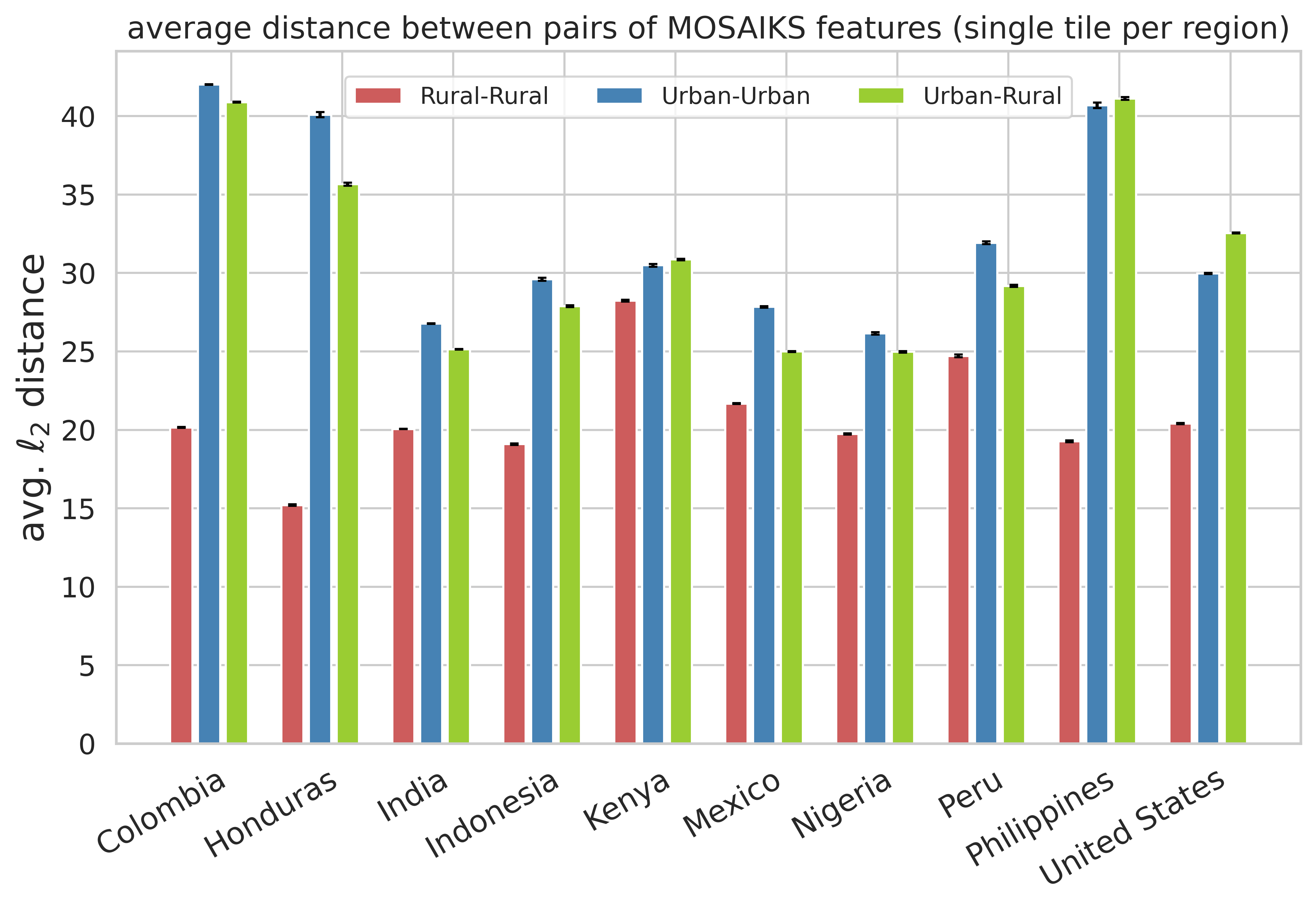}
\end{center}
\caption{Average distance $\ell_2$ between MOSAIKS \emph{tiles} sampled from of pairs of rural instances, pairs of urban instances, and pairs of urban-rural instances. 
For India, we randomly sub-sample 2,000 rural and 2,000 urban regions to estimate average distances. 
}
\label{fig:feature_distances_1_tile_per_region}
\end{figure}

To understand the amount to which the averaging of the $1 \times 1$km  tiles in the per-region MOSAIKS feature representation affects the distances plotted in  \ref{representation_figure}A, in Figure \ref{fig:feature_distances_1_tile_per_region} we plot distances between the RCF featurizations of individual tiles in each region, where one tile is sampled for each region. 
As in Figure \ref{representation_figure}A, the tile feature distances in  Figure \ref{fig:feature_distances_1_tile_per_region} are smaller between pairs of rural instances than between pairs of urban instances or pairs of one rural and one urban instance.
For all types of pairs, the distances between tiles in Figure \ref{fig:feature_distances_1_tile_per_region} tend to be larger than the distances between average feature  representations in Figure \ref{representation_figure}A. This is expected, since averaging many tiles reduces the variation in the feature representation for each region. The averaging generally reduces the average rural-rural and urban-urban distances more than distances between urban-rural pairs. This is consistent with the observation that the average feature representations for urban and rural regions are substantially separated for many countries, reflected also in the PCA distribution plots in Figure \ref{representation_figure}B.

\begin{table}[H]
\footnotesize
\begin{tabular}{p{1.5cm}p{2cm}p{3cm}p{1.8cm}p{1.5cm}p{1.5cm}p{1.0cm}p{1.3cm}}
\toprule
\multirow{2}{1.5cm}{}    & \multirow{2}{2cm}{\textbf{Dataset}} & \multirow{2}{3cm}{\textbf{Definition    of regions}} & \multirow{2}{2cm}{\textbf{Number    of Regions}} & \multirow{2}{1.5cm}{\textbf{\%    Rural Regions}} & \multicolumn{3}{l}{\textbf{Tiles    Per Region}}    \\ \cmidrule(l){6-8} 
                     &                                   &                                                    &                                                &                                               & \textit{Minimum} & \textit{Mean} & \textit{Maximum} \\ \midrule
\textbf{Colombia}    & 2010 DHS                          & Clusters                                           & 4,868                                         & 30.1\%                                        & 16               & 52            & 88               \\
\textbf{Honduras}    & 2011 DHS                          & Clusters                                           & 1,128                                          & 56.2\%                                        & 16               & 58            & 86               \\
\textbf{India}          & 2012 SECC                         & Aggregated shrid2s                                             & 63,356                                         & 94.4\%                                        & 0               & 49            &  100              \\
\textbf{Indonesia}   & 2017 DHS                          & Clusters                                           & 1,319                                          & 57.8\%                                        & 16               & 58            & 87               \\
\textbf{Kenya}       & 2014 DHS                          & Clusters                                           & 1,585                                          & 61.2\%                                        & 16               & 60            & 86               \\
\textbf{Mexico}      & 2015 survey               & Municipalities                                     & 2,446                                          & 56.1\%                                        & 6               & 89           & 100              \\
\textbf{Nigeria}     & 2018 DHS                          & Clusters                                           & 1,359                                          & 58.8\%    & 16               & 56            & 86               \\
\textbf{Peru}        & 2012 DHS                          & Clusters                                           & 1,131                                          & 38.8\%                                        & 16               & 47            & 85               \\
\textbf{Philippines} & 2017 DHS                          & Clusters                                           & 1,213                                          & 64.0\%                                        & 16               & 62            & 88               \\
\textbf{US}          & 2019 ACS                          & PUMAs                                              & 2,331                                          & 41.8\%                                        & 12               & 94            & 100              \\  \bottomrule


\end{tabular}
\caption{Summary of datasets.}
\label{table:datasets}
\end{table}



\clearpage

\begin{figure}
\begin{center}
\includegraphics[width=.8\textwidth]{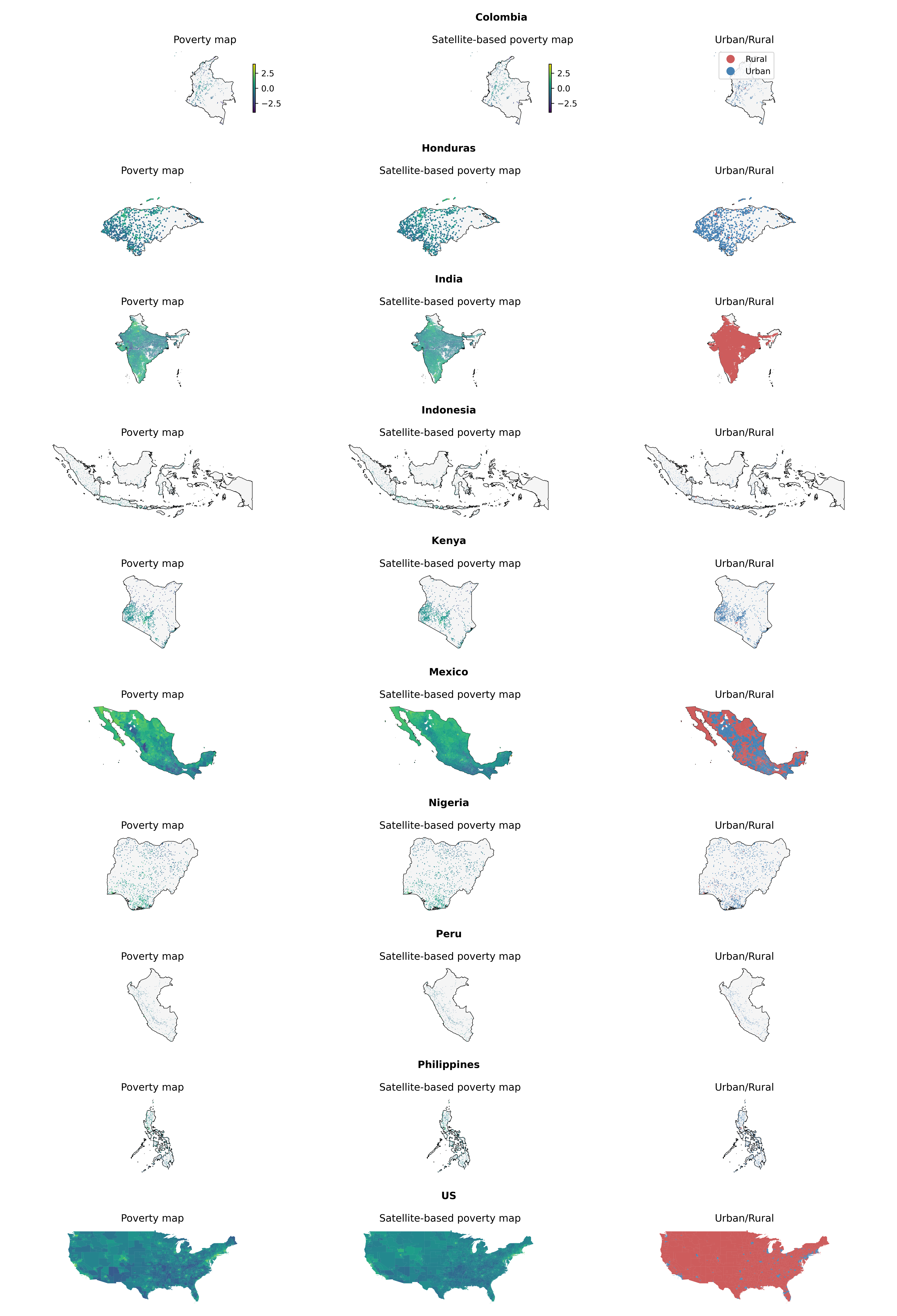}
\end{center}
\caption{Maps of ground truth wealth and satellite-based wealth predictions, as well as categorizations of urban/rural, in each country we study.}
\label{figure:maps}
\end{figure}

%% file: supplement_results.tex
\subsection{Quantifying summary statistics for drivers of allocative unfairness}
In Section \ref{sec:results}, we discussed two main phenomena that could drive allocative unfairness:
\begin{enumerate}
    \item A reversion towards the sample mean, which biases predictions wealth of rural places to be higher than the true value, and predictions of wealth of urban places to be lower, on average, and 
    \item A potential to miss the urban poor, due in part to relying on correlations between urbanization and wealth to produce poverty predictions.
\end{enumerate}

To study these two drivers more rigorously, we evaluate the correlation between summary statistics representing each of the two phenomena and the difference in allocation in \autoref{figure2b} across 100 simulation runs with different random data splits. 
As a summary statistic for the first phenomenon – reversion to the sample mean – we use the difference in standard deviation between satellite-predicted and true wealth distributions (we will notate this summary statistic as $p_1$). As a summary statistic for the second phenomenon --- reliance on correlations between urbanization and wealth --- we use the Spearman's rank correlation between wealth predictions and predictions for being urban (we will notate this summary statistic as $p_2$). 

To measure differences in allocations, we experiment with two summary statistics for allocational disparities: (a) the difference in allocations (between a targeting method that uses ground truth poverty data and a targeting method that uses satellite-based predictions) to rural areas at a 20\% selection threshold, as shown in \autoref{figure2b} and (b)  the difference in area under the curves in \autoref{figure:allocation}, which summarizes the difference in allocations at \emph{all possible} thresholds. For both these summary statistics, we will notate the allocation to rural areas using ground truth poverty data as $b$, the allocation to rural areas using satellite-based poverty predictions as $\hat{b}$, and the difference between the two as $\hat{b} - b$. 
We find that, across countries, in simulations where the first driver is dominant (that is, reversion to the sample mean plays a key role --- as measured by a large reduction in standard deviation), aid is under-allocated to rural areas. In simulations where the second driver is dominant (that is, the correlation between predictions of wealth and predictions of urban build-up is high), aid tends to be over-allocated to rural areas (\autoref{pcorrelations}).

\begin{table}[H]
\footnotesize
\centering
\begin{tabular}{@{}llllll@{}}
\toprule
Country     & (A) $\hat{b}$ & (B) $b$ & (C) $\hat{b} - b$ & (D) Pearson's $r(\hat{b} - b, p_1)$ & (E) Pearson's $r(\hat{b} - b, p_2)$ \\ \midrule
\multicolumn{6}{l}{\textit{Panel   A: Using allocations at a 20\% threshold as the summary statistic for allocations}}                      \\
\textbf{Colombia}    & 84.684            & 88.918         & -4.234***             & -0.026                      & 0.105                     \\
\textbf{Honduras}    & 96.263            & 99.719         & -3.456***             & -0.097                      & -0.094                    \\
\textbf{India}      & 99.290           & 99.972       & -0.682***             & -0.091                     & 0.115                 \\
\textbf{Indonesia}   & 96.515            & 97.818         & -1.303***             & -0.124                     & -0.056                     \\
\textbf{Kenya}       & 87.612            & 90.888         & -3.275***             & -0.306                     &  0.014                      \\
\textbf{Mexico}      & 88.496            & 90.154         & -1.659***             & -0.100                     & 0.111                     \\
\textbf{Nigeria}     & 90.652            & 93.072         & -2.420***             & -0.047                     & -0.110                   \\
\textbf{Peru}        & 92.439            & 97.351         & -4.912***	             & -0.126	                  & 0.042                     \\ 
\textbf{Philippines} & 93.623            & 90.279         & 3.344***              & -0.057                     & 0.193                     \\
\textbf{US}          & 52.077            & 50.000         & ~2.077***              & -0.110                     & 0.469                   \\ \\
\multicolumn{6}{l}{\textit{Panel   B: Using area under the targeting curves (Figure \ref{figure:allocation} Panel D) as the summary statistic for allocations}}                                                  \\
\textbf{Colombia}    & 0.572             & 0.594          & -0.022***             & ~0.063                      & 0.044                    \\
\textbf{Honduras}    & 0.816             & 0.827          & -0.010***             & -0.015                     & 0.107                   \\
\textbf{India}      & 0.953            & 0.961         &    -0.007***         &    -0.090                   & -0.032                      \\
\textbf{Indonesia}   & 0.821             & 0.826          & -0.005***             & -0.046                     & 0.129                     \\
\textbf{Kenya}       & 0.781             & 0.804          & -0.023***             & -0.273                     & -0.012                    \\
\textbf{Mexico}      & 0.718             & 0.745          &       -0.027***       &     ~0.124               & 0.185                 \\
\textbf{Nigeria}     & 0.783             & 0.790          & -0.007***             & -0.049                     & -0.070                     \\
\textbf{Peru}        & 0.682             & 0.696          & -0.015***             & -0.159                     & 0.169                     \\ 
\textbf{Philippines} & 0.834             & 0.804          & 0.030***              & -0.154                     & 0.187                     \\
\textbf{US}          & 0.500             & 0.476          & ~0.024***              & -0.167                     & 0.488                      \\ \bottomrule
\end{tabular}
\caption{Drivers of allocative unfairness between urban and rural areas. Columns A-B compare allocations when using true ($b$) and predicted ($\hat{b}$) values, averaged across runs. Column C documents the average difference in allocations ($b - \hat{b}$), with statistical significance determined via a two-sided t test. Column D records the correlation between our summary statistic for the first driver of allocative unfairness ($p_1$, the magnitude of the gap in standard deviation between true and predicted values) and the difference in allocation to rural areas, across runs. A negative correlation indicates that in general, on runs where the first driver is strong, aid tends to be under-allocated to rural areas. Column E records the correlation between our summary statistic for the second driver ($p_2$, the rank correlation between predicted poverty and predicted urbanization) and the difference in allocation to rural areas, across runs. A positive correlation indicates that in general, on runs where the first driver is strong, aid tends to be over-allocated to rural areas.}
\label{pcorrelations}
\end{table}


\subsection{Additional tables and figures}

\begin{table}[H]
\footnotesize
\begin{tabular}
{p{1.5cm}p{1.5cm}p{1.5cm}p{1.5cm}p{1.5cm}p{1.5cm}p{1.5cm}p{1.5cm}p{1.5cm}}
\toprule
                     & \multicolumn{3}{L{4.5cm}}{\textbf{(A) Predicting poverty}} &  \multicolumn{1}{L{1.5cm}}{\textbf{(B) Predicting urban}} & \multicolumn{2}{L{3cm}}{\textbf{(C) Relating poverty and urban build-up}} & \multicolumn{2}{L{3cm}}{\textbf{(D) Using urban predictions to measure poverty}} \\ \midrule
                     & $R^2(w, \hat{w})$ ~ & Pearson's $r(w, \hat{w})$  & Spearman's $\rho(w, \hat{w})$ & AUC$(u, \hat{u})$                                     & Pearson's $r(w, u)$             & Spearman's $\rho(w, u)$       & Pearson's $r(w, \hat{u})$               & Spearman's $\rho(w, \hat{u})$)              \\ \midrule
\textbf{Colombia}    & 0.70 (0.01)  & 0.84 (0.01)             & 0.83 (0.01)             & 0.94 (0.01)                                      & 0.77 (0.01)                      & 0.72 (0.01)                       & 0.71 (0.02)                          & 0.70 (0.02)                          \\
\textbf{Honduras}    & 0.66 (0.04)  & 0.82 (0.02)             & 0.82 (0.02)             & 0.95 (0.01)                                      & 0.76 (0.03)                      & 0.75 (0.02)                       & 0.77 (0.02)                          & 0.75 (0.02)                          \\
\textbf{India} & 0.52 (0.01)    &  0.72 (0.00) &  0.74 (0.00)  & 0.84 (0.01)        &      0.35 (0.01)       &  0.30 (0.01)                                   &          0.28 (0.01)                            &      0.26 (0.01)            \\
\textbf{Indonesia}   & 0.58 (0.03)  & 0.77 (0.02)             & 0.78 (0.02)             & 0.93 (0.02)                                      & 0.72 (0.02)                      & 0.72 (0.02)                       & 0.72 (0.03)                          & 0.71 (0.03)                          \\
\textbf{Kenya}       & 0.58 (0.03)  & 0.77 (0.02)             & 0.77 (0.02)             & 0.84 (0.02)                                      & 0.59 (0.02)                      & 0.60 (0.03)                       & 0.58 (0.03)                          & 0.55 (0.04)                          \\
\textbf{Mexico}      & 0.66 (0.03)  & 0.82 (0.01)             & 0.83 (0.01)             & 0.79 (0.02)                                      & 0.51 (0.03)                      & 0.51 (0.03)                       & 0.56 (0.04)                          & 0.57 (0.04)                          \\
\textbf{Nigeria}     & 0.65 (0.02)  & 0.81 (0.01)             & 0.83 (0.01)             & 0.87 (0.02)                                      & 0.57 (0.03)                      & 0.58 (0.03)                       & 0.71 (0.02)                          & 0.72 (0.03)                          \\
\textbf{Peru}        & 0.69 (0.03)  & 0.83 (0.02)             & 0.83 (0.02)             & 0.96 (0.01)                                      & 0.77 (0.02)                      & 0.77 (0.02)                       & 0.77 (0.02)                          & 0.74 (0.02)                          \\
\textbf{Philippines} & 0.47 (0.09)  & 0.70 (0.04)             & 0.72 (0.03)             & 0.90 (0.02)                                      & 0.53 (0.03)                      & 0.53 (0.03)                       & 0.62 (0.03)                          & 0.63 (0.03)                          \\
\textbf{US}          & 0.49 (0.05)  & 0.72 (0.03)             & 0.71 (0.02)             & 0.99 (0.00)                                      & 0.28 (0.04)                      & 0.28 (0.04)                       & 0.28 (0.04)                          & 0.24 (0.05)                          \\  \bottomrule

\end{tabular}
\caption{Relationship between urban build-up and predicting wealth from satellite imagery. Panel A evaluates the predictive accuracy of our satellite-based wealth predictions using three metrics ($R^2$, Pearson's $r$, and Spearman's $\rho$). Panel B evaluates the predictive accuracy of our satellite-based urban/rural classifications based on AUC. Panel C records the correlation between wealth and an indicator variable for being urban in each country (using Pearson's $r$ and Spearman's $\rho$). Panel D records the correlation between wealth and a satellite-based prediction of being urban in each country (using Pearson and Spearman $\rho$). $w$ represents ground-truth wealth; $\hat{w}$ predicted wealth; $u$ ground-truth urban (a binary indicator), and $\hat{u}$ predicted urban (a probabilistic prediction between 0 and 1). Standard deviations across bootstrapped runs are shown in parentheses.}
\label{table:urban}
\end{table}

\begin{figure}[H]
\begin{center}
\includegraphics[width=.8\textwidth]{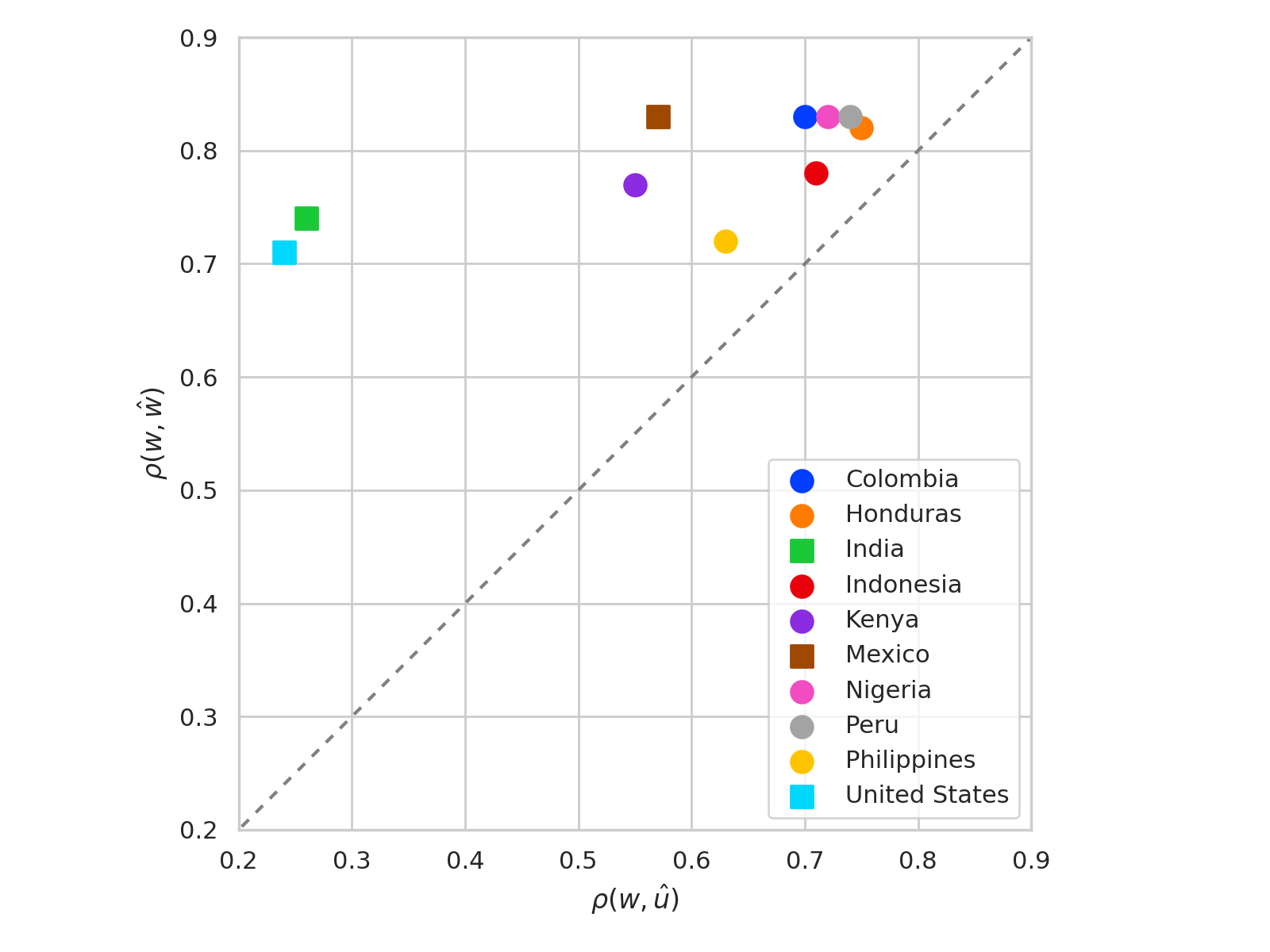}
\end{center}
\caption{Comparing the predictive accuracy (measured with Spearman's $\rho$ of satellite-based poverty predictions ($\hat{w})$ for identifying wealth ($w$), in comparison to using satellite-based probabilistic predictions of being urban ($\hat{u}$) for identifying wealth ($w$).}
\label{figure:urbanwealth}
\end{figure}


 
\begin{figure}[H]
\begin{center}
\includegraphics[width=.6\textwidth]{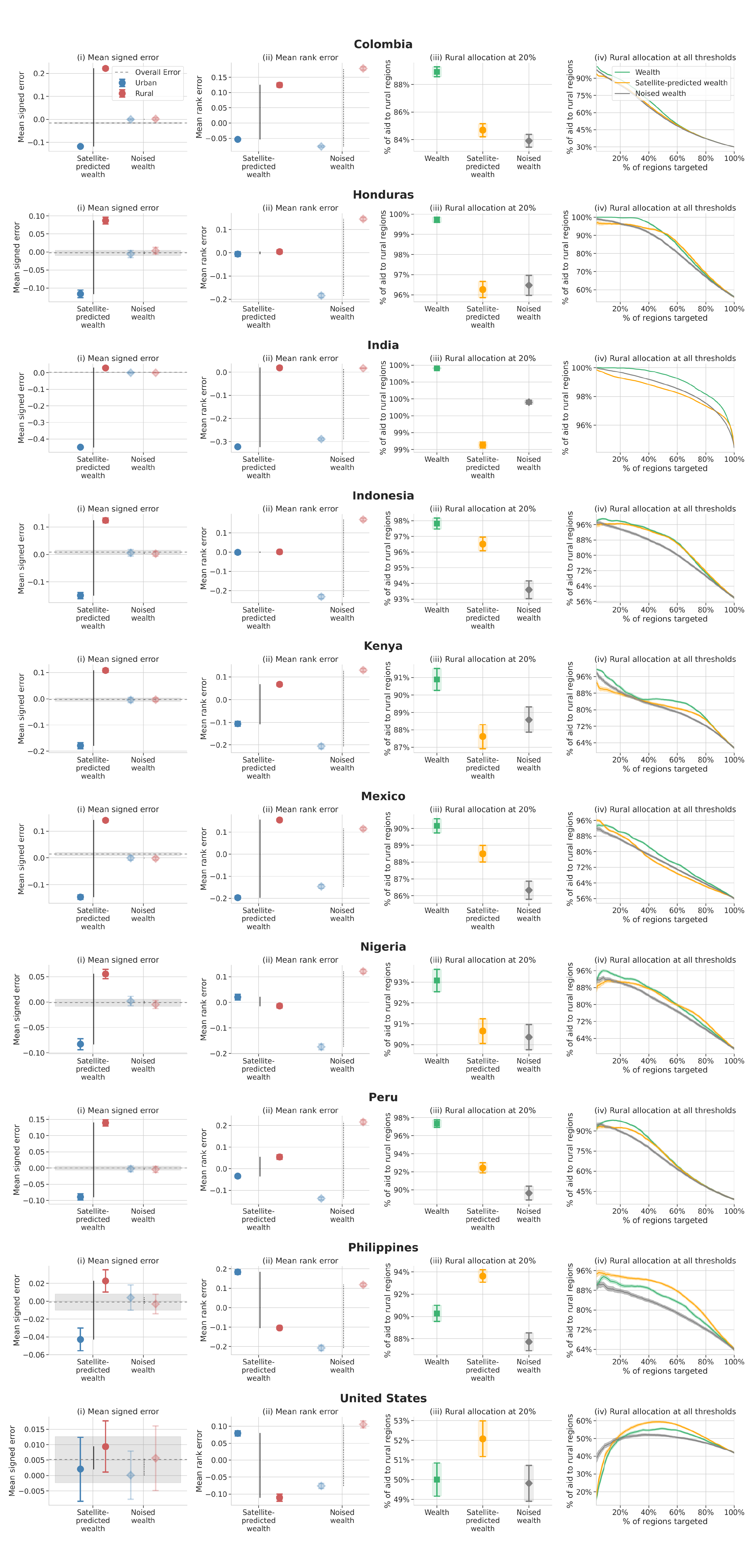}
\end{center}
\caption{Allocative bias in using satellite-based wealth estimates. Panel A compares the mean signed error for satellite-based wealth predictions (left) to the noised-wealth baseline (right). Panel B makes the same comparison for the mean rank error. Panel C records the share of rural regions targeted in a hypothetical aid program targeting the poorest 20\% of regions in each country, depending whether ground-truth (green) wealth, satellite-based wealth estimates (yellow) or the noised-wealth baseline (gray) are used. Panel D records the sensitivity of the allocations from Panel C to the eligibility threshold. In all panels error bars represent two standard errors above and below the mean.}
\label{figure:allocation}
\end{figure}


\begin{figure}[H]
\includegraphics[width=\textwidth]{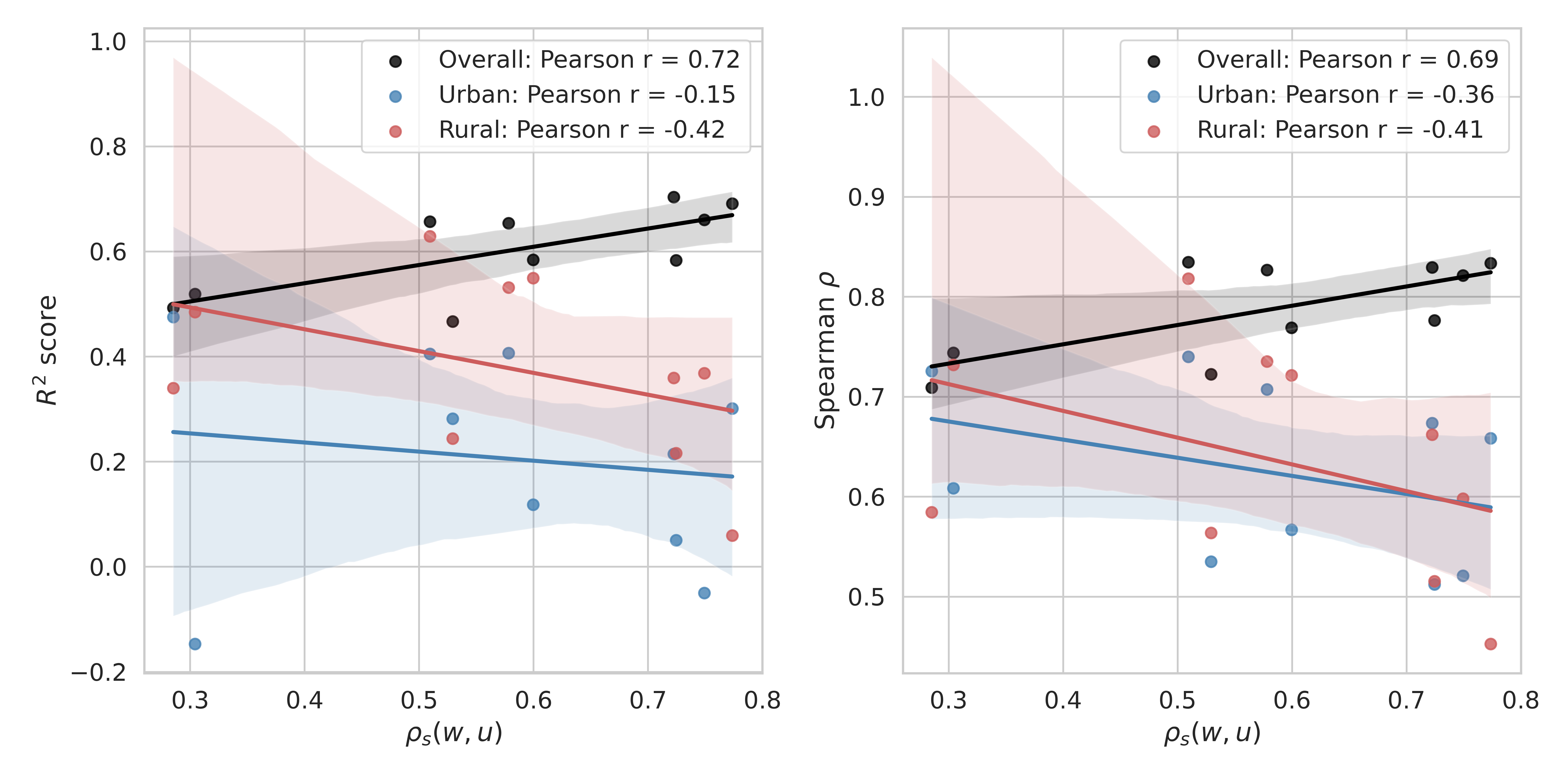}
\caption{Predicted performance ($R^2$ score and Spearman $\rho$) vs. the degree of rank correlation between wealth and binary urbanization values (urban or rural). Colors represent the evaluation regime: overall performance (black), performance across only rural regions (red), and performance across only urban regions (blue). Each dot represents one evaluation regime for one country.}
\label{figure:correlation_predicts_performance}
\end{figure}

\begin{figure}[H]
\includegraphics[width=\textwidth]{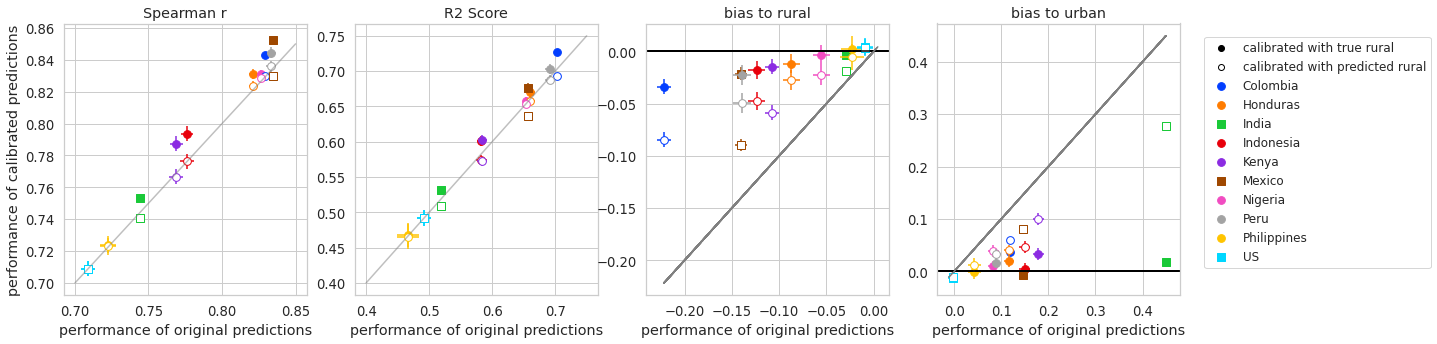}
\caption{Additive recalibration by group raises the linear fit and rank correlation of overall predictions (leftmost two panels) and  reduces statistical bias of predictions per-group (rightmost two panels).}
\label{figures1}
\end{figure}